\documentclass[sigconf,10pt]{acmart}
\renewcommand\footnotetextcopyrightpermission[1]{}
\AtBeginDocument{%
  \providecommand\BibTeX{{%
    \normalfont B\kern-0.5em{\scshape i\kern-0.25em b}\kern-0.8em\TeX}}}

\usepackage{array}
\usepackage{balance}
\usepackage{stfloats}
\usepackage{soul}
\usepackage[italic]{mathastext}
\usepackage{graphicx}
\usepackage{textcomp}
\usepackage{xcolor}
\usepackage[shortlabels]{enumitem}
\usepackage{tcolorbox}
\usepackage{xspace}
\usepackage{url}
\usepackage{booktabs}
\usepackage{amsmath}
\usepackage[capitalise]{cleveref}
\usepackage{algorithm}
\usepackage{caption}
\usepackage{subcaption}
\usepackage{gensymb}
\usepackage[noend]{algpseudocode}
\usepackage{wrapfig}
\usepackage{hyperref}
\usepackage{upquote}
\usepackage{listings}

\definecolor{mygray}{HTML}{d6d0d6}

\newcommand{\etal}{\emph{et al. }}

\newcommand{\name}{\texttt{Steward}\xspace}
\newcommand{\nametitle}{\textsc{Steward}\xspace}

\newtheorem{thm}{Theorem}[section] 

\newcommand{\thistheoremname}{}
\newtheorem{genericthm}[thm]{\thistheoremname}

\definecolor{lightgray}{rgb}{0.95, 0.95, 0.95}
\definecolor{darkgray}{rgb}{0.4, 0.4, 0.4}
\definecolor{editorGray}{rgb}{0.95, 0.95, 0.95}
\definecolor{editorOcher}{rgb}{1, 0.5, 0} 
\definecolor{editorGreen}{rgb}{0, 0.5, 0} 
\definecolor{orange}{rgb}{1,0.45,0.13}		
\definecolor{olive}{rgb}{0.17,0.59,0.20}
\definecolor{brown}{rgb}{0.69,0.31,0.31}
\definecolor{purple}{rgb}{0.38,0.18,0.81}
\definecolor{lightblue}{rgb}{0.1,0.57,0.7}
\definecolor{lightred}{rgb}{1,0.4,0.5}

\lstdefinelanguage{CSS}{
  keywords={color,background-image:,margin,padding,font,weight,display,position,top,left,right,bottom,list,style,border,size,white,space,min,width, transition:, transform:, transition-property, transition-duration, transition-timing-function},	
  sensitive=true,
  morecomment=[l]{//},
  morecomment=[s]{/*}{*/},
  morestring=[b]',
  morestring=[b]",
  alsoletter={:},
  alsodigit={-}
}

\lstdefinelanguage{JavaScript}{
  morekeywords={typeof, new, true, false, catch, function, return, null, catch, switch, var, if, in, while, do, else, case, break},
  morecomment=[s]{/*}{*/},
  morecomment=[l]//,
  morestring=[b]",
  morestring=[b]'
}

\lstdefinelanguage{HTML5}{
  language=html,
  sensitive=true,	
  alsoletter={<>=-},	
  morecomment=[s]{<!-}{-->},
  tag=[s],
  otherkeywords={
  >,
	<!DOCTYPE,
  </html, <html, <head, <title, </title, <style, </style, <link, </head, <meta, />,
	</body, <body,
	</div, <div, </div>, 
	</p, <p, </p>,
	</script, <script,
    </a, <a,
    <img,
    <li, </li,
  <canvas, /canvas>, <svg, <rect, <animateTransform, </rect>, </svg>, <video, <source, <iframe, </iframe>, </video>, <image, </image>, <header, </header, <article, </article
  },
  ndkeywords={
  =,
  charset=, src=, id=, width=, height=, style=, type=, rel=, href=, alt=, component-name=, class=, aria-label=, data-testid=,
  fill=, attributeName=, begin=, dur=, from=, to=, poster=, controls=, x=, y=, repeatCount=, xlink:href=,
  margin:, padding:, background-image:, border:, top:, left:, position:, width:, height:, margin-top:, margin-bottom:, font-size:, line-height:,
  transform:, -moz-transform:, -webkit-transform:,
  animation:, -webkit-animation:,
  transition:,  transition-duration:, transition-property:, transition-timing-function:,
  }
}

\lstdefinestyle{htmlcssjs} {%
  basicstyle={\footnotesize\ttfamily},   
  frame=b,
  xleftmargin={0.75cm},
  numbers=left,
  stepnumber=1,
  firstnumber=1,
  numberfirstline=true,	
  identifierstyle=\color{black},
  keywordstyle=\color{blue}\bfseries,
  ndkeywordstyle=\color{editorGreen}\bfseries,
  stringstyle=\color{editorOcher}\ttfamily,
  commentstyle=\color{brown}\ttfamily,
  language=HTML5,
  alsolanguage=JavaScript,
  alsodigit={.:;},	
  tabsize=2,
  showtabs=false,
  showspaces=false,
  showstringspaces=false,
  extendedchars=true,
  breaklines=true,
  literate=%
  {Ö}{{\"O}}1
  {Ä}{{\"A}}1
  {Ü}{{\"U}}1
  {ß}{{\ss}}1
  {ü}{{\"u}}1
  {ä}{{\"a}}1
  {ö}{{\"o}}1
}
\lstdefinestyle{py} {%
language=python,
literate=%
*{0}{{{\color{lightred}0}}}1
{1}{{{\color{lightred}1}}}1
{2}{{{\color{lightred}2}}}1
{3}{{{\color{lightred}3}}}1
{4}{{{\color{lightred}4}}}1
{5}{{{\color{lightred}5}}}1
{6}{{{\color{lightred}6}}}1
{7}{{{\color{lightred}7}}}1
{8}{{{\color{lightred}8}}}1
{9}{{{\color{lightred}9}}}1,
basicstyle=\footnotesize\ttfamily, 
numbers=left,               
numbersep=5pt,              
tabsize=4,                  
extendedchars=true,         %
breaklines=true,            
keywordstyle=\color{blue}\bfseries,
frame=b,
commentstyle=\color{brown}\itshape,
stringstyle=\color{editorOcher}\ttfamily, 
showspaces=false,           
showtabs=false,             
xleftmargin=17pt,
framexleftmargin=17pt,
framexrightmargin=5pt,
framexbottommargin=4pt,
showstringspaces=false,      
}%
\makeatother

\begin{abstract}
    Recently, large language models (LLMs) have demonstrated exceptional capabilities in serving as the foundation for AI assistants. One emerging application of LLMs, navigating through websites and interacting with UI elements across various web pages, remains somewhat underexplored. We introduce \name, a novel LLM-powered web automation tool designed to serve as a cost-effective, scalable, end-to-end solution for automating web interactions. Traditional browser automation frameworks like Selenium, Puppeteer, and Playwright are not scalable for extensive web interaction tasks, such as studying recommendation algorithms on platforms like YouTube and Twitter. These frameworks require manual coding of interactions, limiting their utility in large-scale or dynamic contexts. \name addresses these limitations by integrating LLM capabilities with browser automation, allowing for natural language-driven interaction with websites. \name operates by receiving natural language instructions and reactively planning and executing a sequence of actions on websites, looping until completion, making it a practical tool for developers and researchers to use. It achieves high efficiency, completing actions in 8.52 to 10.14 seconds at a cost of \$0.028 per action or an average of \$0.18 per task, which is further reduced to 4.8 seconds and \$0.022 through a caching mechanism. It runs tasks on \textit{real websites} with a 40\% completion success rate. We discuss various design and implementation challenges, including state representation, action sequence selection, system responsiveness, detecting task completion, and caching implementation. 

\end{abstract}

\begin{document}

\title{
\nametitle: Natural Language Web Automation}

\author{Brian Tang}
\email{bjaytang@umich.edu}
\affiliation{
  \institution{University of Michigan}
  \city{Ann Arbor}
  \country{MI, USA}}

\author{Kang G. Shin}
\email{kgshin@umich.edu}
\affiliation{
  \institution{University of Michigan}
  \city{Ann Arbor}
  \country{MI, USA}}


\settopmatter{printfolios=true}
\maketitle
\pagestyle{plain}

\section{Introduction} \label{sec:introduction}

Simulating user navigation and interactions on the web 
is required for a number of use-cases, including web 
measurement studies, UI testing and debugging, 
analyzing privacy practices, etc. 
The state-of-the-art (SOTA) approaches require the use of 
a browser automation framework like Selenium, Puppeteer, 
or Playwright to manually trace, record, and code 
interactions with HTML elements. 
This process is infeasible for conducting large-scale tests with 
many webpage contexts or multiple websites. Consider the task of 
studying recommendation algorithm behavior. 
With such content being dynamically generated and 
location/context-dependent, relying solely on a browser 
automation tool to record and playback actions would not scale.

Recently, large language models (LLMs) have demonstrated 
exceptional capabilities in serving as the foundation for
AI assistants. They have been used widely for aiding users 
in a variety of applications, 
including text writing, task assistance, reasoning, information 
retrieval, Q\&A, code interpretation, translation, and so on. 
Numerous assistant services built on LLMs have sprung up 
in the wake of ChatGPT~\cite{ouyang2022instructgpt}, 
an instruction-tuned LLM created by OpenAI.

\begin{figure}[t]
    \centering
    \includegraphics[width=.98\columnwidth]{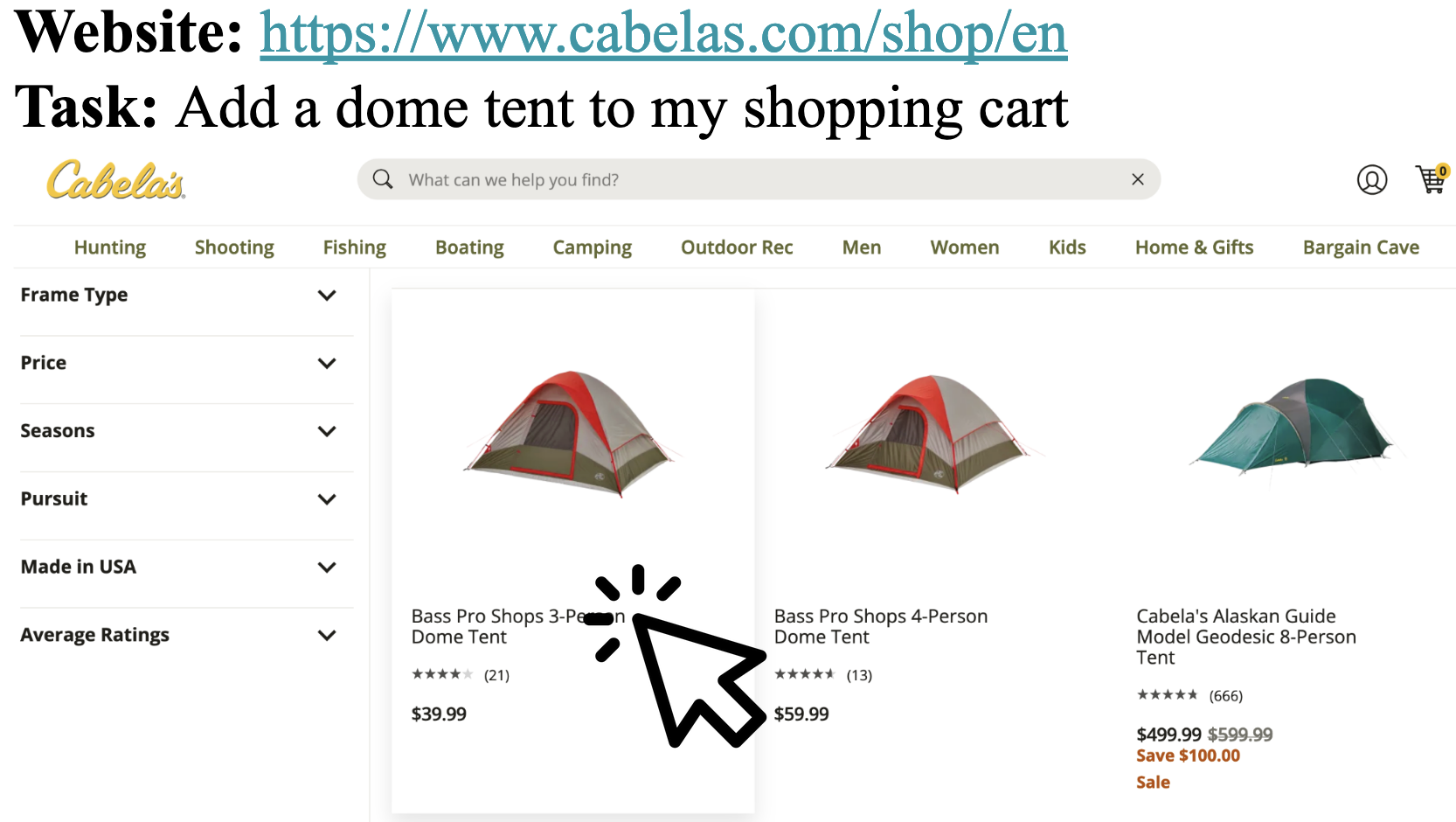}
    \caption{Given a natural language task, \name iteratively selects UI elements to perform actions and interact with. \name performs the actions in a browser automation tool.}
    \label{fig:teaser}
\end{figure}

However, their (in)ability to interact with websites across 
various contexts and web pages remains under-explored despite 
its importance. LLMs are effective in predicting future 
states and tokens based on the currently provided context, 
a capability that naturally extends itself to performing 
activities on the web. So far, AI assistants have been 
limited to using search engine APIs, visiting URLs, and 
scraping site content for information retrieval. 
Very few systems have been created to enable these 
language models to \textit{interact} with websites. 
Due to the complexity of content on the Internet, the SOTA 
approaches to AI services, such as ChatGPT, have focused on 
integrating developer-made APIs for popular online services and 
platforms~\cite{chatgptapi}. These approaches, while more 
consistent and robust, are less scalable than the 
alternative --- granting LLMs the ability to interface 
with a browser automation tool.

Augmenting LLMs with web capabilities is a challenging task that 
could reap numerous benefits by providing ad hoc intelligent crawling or API services to 
users, developers, and researchers. This tool
could provide flexible assistance vastly beyond the capabilities 
of current and prior assistants. 
With careful prompt engineering, fine-tuning, and system 
design, language models can perform complex sequences of 
actions and activities on the Internet on behalf of users. 

\subsection{\name}

We propose \name, a fast, reliable, and cost-efficient 
LLM-powered web automation tool. It is an end-to-end system 
that takes a natural language instruction as input and 
performs operations on websites until the end state is 
detected/reached. 
\name can simulate user behaviors on websites and even 
perform entire tasks to completion on real websites, 
for example, adding items to e-commerce site shopping 
carts, searching for and sharing YouTube videos, 
booking tickets, checking flight/lodging status or 
availability, etc. Using OpenAI's language and vision 
model APIs, \name intelligently and reactively 
performs actions on sites in only 8.52--10.14 seconds and 
at a cost of \$0.028 per action. \name also uses a webpage 
interaction caching mechanism that 
reduces runtime to 4.8 seconds and \$0.013 per action.

\name is also generalizable to handle previously unseen 
web contexts and perform correct action sequences even 
after sites update or remove their content. 
Rather than relying on fine-tuning or training on 
datasets, \name uses purely zero/few-shot prompting. 
Thus, it is easily deployable, scalable, and relatively 
low-cost, allowing for plug-and-play integration 
with any LLM.\footnote{
\url{https://github.com/byron123t/Steward}}.

\subsection{Technical Challenges}

Determining the correct action--element pair to perform on 
a website requires careful design of a state representation 
of the webpage's current context. Selecting a correct 
action sequence on a site using minimal input/output tokens
while maintaining accuracy poses another significant 
challenge. Finally, implementing a system design that runs 
in the order of seconds constitutes the last major challenge.

The SOTA systems have limitations, such as using proprietary 
models, lacking in reliability and scalability, cost more, 
or are not end-to-end systems. They mainly lack integration
with browser automation tools and are impractical for 
larger intelligent web crawls or real-time operation.

\subsection{Our Contributions}

Our work consists of two main thrusts: (1) designing an 
automated framework for modeling the current website 
contexts and executing UI actions, and (2) the analysis 
of \name's performance with respect to various criteria.

\noindent We address each of the above technical challenges 
and make the following main contributions:
\begin{enumerate}[noitemsep,leftmargin=0.4cm,topsep=5pt]
   
  \item A unique LLM-powered web execution procedure that easily 
  fits into browser automation frameworks. \name was designed 
  specifically for use with the Playwright framework. 
  It is fully autonomous and only requires the user to 
  type out a high-level goal/task to perform on a website 
  in natural language.
  
  \item A context-aware, site/app-agnostic UI exercising
  system capable of automating web interactions on a large 
  scale. \name can generalize its knowledge to navigate 
  and interact with a variety of websites.
  For the top 5 elements, \name is capable of achieving a 
  top-1 action + element selection accuracy of 81.44\% 
  without any training or fine-tuning. 
  It achieves a per-step accuracy of 46.70\% on the Mind2Web 
  benchmark and a 40\% task completion rate, able to execute
  roughly 56\% of the tasks' actions until encountering an
  error.

  \item An in-depth evaluation of \name's runtime and cost 
  in various configurations. Its system design is optimized for 
  runtime and cost efficiency, achieving a median runtime of 
  8.52 seconds or 10.14 seconds with text entry at a cost of 
  \$0.028 per action. It also includes an implementation of 
  a caching mechanism for storing and reusing 
  website interactions, which reduces the runtime and cost 
  of a step by 43.7\% and 53.6\%, respectively.

\end{enumerate}

\section{Background} \label{sec:background}

\subsection{AI UI Exercising Tools}

Earlier web UI automation tools have sought to use natural 
language inputs to perform complex user interaction sequences 
on websites. 
Earlier approaches were not as generalizable to 
unseen websites or interaction sequences, either. 
For example, some of the earlier works exploring advanced 
natural language UI exercising that leverage NLP and 
reinforcement learning resulted in lower success rates on 
websites and interactions outside of the training set 
distribution~\cite{li2021glider,mazumder2020flin}. 
By using LLMs and achieving a balance 
between a concise yet rich web page representation, 
\name achieves accurate and broad coverage comparable 
to prior approaches.

Glider~\cite{li2021glider}, an automated and scalable 
generation of web automation scripts (tasklets) 
from a natural language task query and a website URL, uses 
hierarchical reinforcement learning to navigate the 
website's UI and maximize rewards based on task progress. 
It generates tasklets which are sequences of actions 
to perform on a particular site.

FLIN~\cite{mazumder2020flin} proposes a natural language 
interface for web navigation that maps user commands
to concept-level actions. The authors frame their approach 
as a ranking problem: scoring the most relevant navigation 
instruction given a user command and a webpage. 
By using semantic similarity, action keywords, and the 
BERT~\cite{devlin2018bert} model, FLIN is able to perform 
basic high-level tasks.

\subsection{Natural Language UI Testing}

Other earlier related works use natural language generation 
to augment UI testing frameworks by generating labels, 
comments, test inputs, test cases, and more.

CrawLabel~\cite{liu2022crawlabel} introduces techniques 
to compute natural-language labels for end-to-end UI 
test-cases in web applications by using information from 
the browser's document object model (DOM).

The work of Wanwarang \etal~\cite{wanwarang2020testing}
introduces an approach called {\em Link} 
for generating realistic test inputs for mobile apps. 
It leverages knowledge bases and uses label matching, NLP, 
and clustering, to cover more statements than 
randomly-generated text inputs for testing mobile apps.

Kirinuki \etal~\cite{kirinuki2021nlp} propose
script-free testing in web application development, 
using NLP and heuristic search algorithms to identify 
web elements and determine test procedures based on 
test-cases written in a domain-specific language, 
identifying the web elements to be operated.

Deng \etal propose creating a general natural language 
interface for the web to make the Internet more accessible 
to users with disabilities. They present their ongoing 
efforts of curating a benchmark dataset of websites and 
tasks~\cite{deng2023more}.

Humanoid~\cite{li2019humanoid} is a deep learning-based 
approach to generating test inputs for mobile apps by 
learning from human interactions in the RICO \cite{deka2017rico} 
dataset, prioritizing inputs based on 
their perceived importance by users.

\subsection{LLM Web Automation Systems}

Since the advent of large language models, various open-source 
tools and datasets have been created to augment its capabilities. 
For example, AutoGPT~\cite{AutoGPT} was created as an 
autonomous AI agent that continually loops until its 
high-level tasks are achieved. While accessing the Internet 
via BeautifulSoup for web scraping and information retrieval, 
it is unable to perform actions on sites.
Another popular plugin, WebPilot~\cite{webpilot}, 
a ChatGPT plugin, grants access to a Bing search API and 
basic website information retrieval capabilities. 
Unlike these two approaches, Natbot~\cite{natbot} was created 
as an early-stage prototype exploring web interactions 
using OpenAI's Davinci model and few-shot prompting.

\name is not the first to explore the use of LLM's potential 
for web automation; several researchers have achieved 
various levels of success in natural language web automation.

Mind2Web~\cite{deng2023mind2web} is a dataset recently 
created by researchers. The dataset is a
record of Amazon Mechanical Turk users' actions on a web 
browser using the Playwright~\cite{playwright} automation 
framework. It serves as the primary data source for 
evaluating the language models used in our design and our 
overall system (\cref{sec:design}). The authors of
\cite{deng2023mind2web} train their own language 
model using a derivative model of BERT.

Gur \etal performed an analysis of LLM HTML understanding 
using various transformers like T5~\cite{roberts2019t5} and 
LaMDA~\cite{Collins2021lamda}. They subsequently created 
WebAgent~\cite{gur2023real}, another HTML-T5 type language model 
combined with a Flan-U-PaLM~\cite{chung2022scaling} program 
synthesis model trained on CommonCrawl~\cite{commoncrawl} 
HTML and web interaction data. Their specially trained 
540B parameter model achieves the best-known accuracy
on the Mind2Web benchmark dataset.

Multimodalweb agents have also been a direction of interest as 
WebGUM~\cite{furuta2023multimodal} uses T5~\cite{roberts2019t5} 
as an image and embedding encoder and a decoder that selects 
actions and elements. Another more general task automation 
framework leverages GPT-4~\cite{zhang2023responsible} using a 
combination of language and vision models to perform actions 
on websites and the user's OS.

Sodhi \etal~\cite{sodhi2023heap} proposed a contextual 
Markov decision process (MDP) in which the context is the web 
task objective expressed as an instruction, implicitly in a 
conversation, or as a set of structured parameters. The state 
of the MDP is the DOM of the current webapge, and the action and 
transition functions are based on clicking and typing to interact 
with an element (represented as an id and a value string). 
Their system uses hierarchical 
prompting to break down complex tasks into smaller policies. 
They note that there are several limitations, for example, 
when a dropdown menu does not appear in the DOM.

He \etal~\cite{he2024webvoyager} implement their LLM tool by 
leveraging GPT-4-ACT~\cite{gpt4vact}, 
a GPT-4-Vision augmentation 
that uses \textit{set of mark prompting}~\cite{yang2023set}, 
to label and reference webpage elements on a screenshot. 
Their vast majority (as much as 91\%) of 
errors result from navigation getting stuck, the language 
model hallucinating, or an issue with their visual 
grounding approach. They are specifically investigating 
GPT-4-Vision's potential for parsing visual and textual 
information on webpages to use in website navigation.

Hong \etal~\cite{hong2023cogagent} create models leveraging 
OCR and visual grounding with captioning datasets for 
representing GUIs via text. Their results appear to outperform 
smaller language models both prompted and fine-tuned for web 
navigation on the Mind2Web dataset for element selection from 
the top 10 elements, selecting the ground truth element up 
to 62.3\% of the time.

\subsection{LLM Web Automation Limitations and More}
Given their limited performance, most of these autonomous 
agents are yet to be a practical solution for day-to-day 
usage~\cite{huq2023s}.
Performing truncation on the HTML to feed into these language 
model agents was shown to significantly improve the
performance over cases without truncation.

Furota \etal~\cite{furuta2024exposing} extensively study the 
transferability of large multimodal agents(LMAs) to more 
realistic sequential task compositions. They design a new 
test bed, {\tt CompWoB}, with 50 compositional tasks.

\subsection{Design Comparison with Related Work}

In the domain of natural language web automation, various 
methodologies have been employed to enhance web navigation 
accuracy. Many of these prior studies share similar design 
principles. A language model is given access to some state 
representation of a website and must select the appropriate 
HTML element and action given a natural language task 
specified by the user. These designs roughly fall into 
the following categories.

\textbf{HTML Element Proposal and Selection: }
This methodology uses an element proposal followed by the 
selection of an appropriate action using a language 
model~\cite{deng2023mind2web,sodhi2023heap}. 
It is simplistic, straightforward, and efficient, but 
yields over-reliance on the model's performance 
and fine-tuning dataset.

\textbf{Planning and Program Synthesis: }
The planning-based approach~\cite{gur2023real} focuses on 
generating correct subtasks or sub-instructions to automate 
a given task. However, this approach requires high accuracy 
in generating subtasks, i.e., the generator must have rich 
context/data (raw HTML) to make an informed prediction. 
It must also be consistent in generating browser automation code.

\textbf{Multimodal Web Automation: }
In contrast to the other approaches, multimodal web
automation emphasizes using screenshots of the webpage
and identifying UI elements to interact with using
semantic segmentation and/or element object detection
\cite{he2024webvoyager,hong2023cogagent,zhang2023responsible}.

\begin{figure*}[t]
    \centering
    \includegraphics[width=\textwidth]{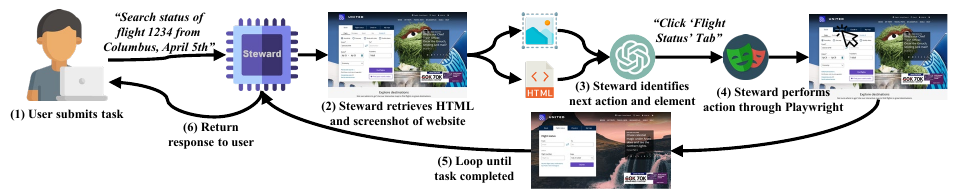}
    \caption{High-level overview of \name performing a task on a website (checking a flight's status on united.com).}
    \label{fig:high_level}
\end{figure*}

\textbf{Our Approach: }
\name adopts an approach most similar to a combination
of the 3 approaches. 
According to our preliminary investigations, this 
yielded fewer errors when used with off-the-shelf models.
Our approach differs from prior work in its
end-to-end system design with a reactive planning-based 
agent. This means our system plans and makes decisions on the 
fly after each step. It is built to work with off-the-shelf 
language/vision models with just zero or few-shot prompting
while operating as quickly and cost-efficiently as possible, 
without sacrificing reliability. It does this by filtering 
and cleaning the set of HTML elements to minimize the amount 
of noise and tokens the language models must process.
Our system also integrates an action caching system to 
further improve runtime and cost-efficiency, reducing the already
low 8 seconds per action step to just 4 seconds per step. 
Finally, our system also accounts for task completion by 
detecting the end state and terminating the program. 
We will discuss the advantages 
of our system and design philosophy in the following section.

\section{Design, Implementation, and Comparison}  
\label{sec:design}

We now detail \name's design and implementation. \name consists of 3 main 
components: (1) a large language model (LLM) and prompting
framework that can handle webpage state representation and
navigation, (2) an HTML cleaner, a runtime/cost-optimized 
execution pipeline, an action caching system, and
(3) an integration with the Playwright browser automation tool.

The design of \name is inspired by how humans perceive, 
process, and interact with websites. The system is built 
to automate tasks that users (i.e., LLM users, website 
developers, and researchers) may want to conduct on websites. 
First, the system analyzes a website, providing a short 
high-level description of the page. 
In parallel with this analysis, the system considers 
its user-provided goal and a screenshot of the page to 
determine the next course of action. This contextual 
information is stored in a state representation that is 
used in every prompt and LLM query. It looks at the 
interactable elements from the DOM (in HTML) to select
an element for the Playwright browser automation tool 
to interact with. 
Finally, the system records the actions taken and 
memorizes these prior action sequences when considering 
the next element to interact with. In addition, 
it caches previously seen contexts and action sequences 
to avoid repetitive calls to the LLMs. \name is built 
as a low-budget intelligent web automation tool that 
minimizes both the runtime and cost of operation.

\begin{figure}[t]
    \centering
    \includegraphics[width=.98\columnwidth]{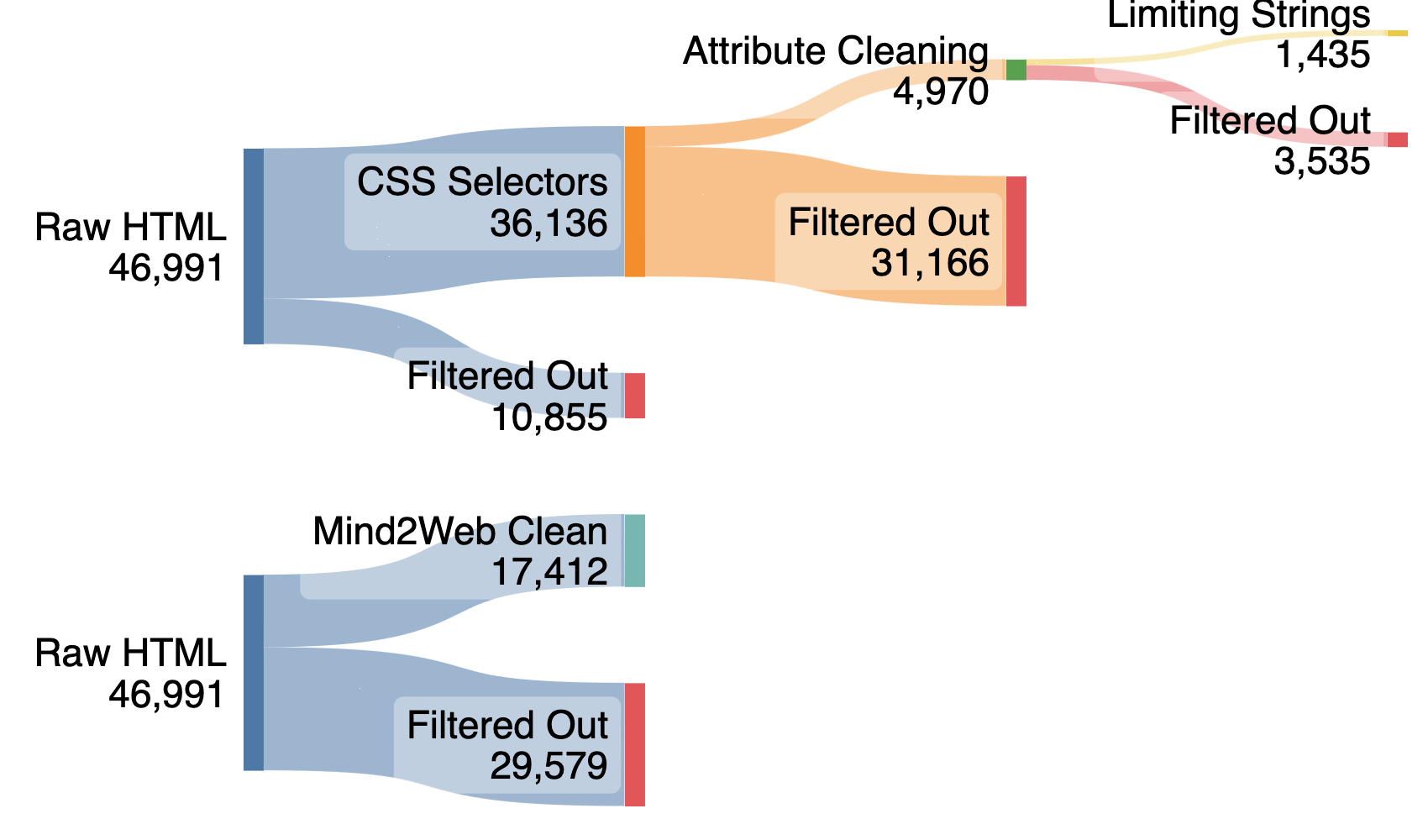}
    \caption{Token Counts for HTML Filtering Approaches. (Top) Our Approach, (Bottom) Deng \etal's Approach~\cite{deng2023mind2web}.}
    \label{fig:html_tokens}
\end{figure}

\subsection{Example: Adding Items to Shopping Cart}

We take the following successfully executed example 
observed in our evaluations with \name. 
For a website like cabelas.com, if \name's current 
user's goal is to ``\textit{Add a dome tent to my 
shopping cart}'', and \name has already performed 
the actions of clicking on ``\textit{Camping}'', 
clicking ``\textit{Close}'' on a promotional popup window, 
and clicking ``\textit{Dome Tents}'', the system will 
``\textit{click on the first dome tent product displayed}''. 
This is because the list of prior actions, the screenshot 
of the web page, and the generated page context have all 
been updated with the change in the web page's state. 
\name will proceed to filter the set of interactable 
web page elements down to a list of 15 elements 
and then select the best match:

\textsc{``click''} \textit{<a href="/shop/en/cabelas-west-wind-4-person-dome-tent" title="cabela's west wind 4-person dome tent"></a>}

After executing this command, the web page will update to this 
4-person dome tent's product page. \name's internal state 
representation of the website will also update, causing it to 
select the best matching element after filtering again:

\textsc{``click''} \textit{<a id="add2cartbtn" href="" role="button" class="test primary pdp\_non\_chart\_addtocart" title="add to cart">add to cart</a>}

\textbf{Constructing Website State: }
After a web page initially loads, or after an action is 
selected and performed on that website, \name first 
``perceives'' the web page by taking a screenshot of the page. 
A vision transformer is prompted to identify the next best 
action to perform to achieve the user's task, and the 
screenshot image is input along with relevant state 
information (website URL and prior actions performed),
e.g., SCREENSHOT RESPONSE: ``click the ``Camping'' 
category on the navigation bar''.

In parallel, the page's plaintext is retrieved from 
the HTML, and a language model is prompted to return 
a brief high-level summary of the current page context,
e.g., PAGE CONTEXT: ``Ecommerce website page for Cabela's 
featuring a variety of tents for camping and outdoor 
activities, with filtering options by brand, type, size, 
and sleeping capacity.''

\subsection{HTML Processing}

The first major challenge with LLM web automation is enabling 
LLMs to parse large HTML representations of websites. 
These HTML snippets must fit within the context lengths 
of the language models while avoiding overcrowding the input. 
To minimize the amount of irrelevant information contained 
in HTML tags, we utilize a three-step approach to filtering 
and cleaning the list of HTML elements on a page.

\textbf{Filtering for Interactable Elements: }
First, \name filters the set of elements to interact with 
using CSS selectors. We are only concerned with 
interactable elements, which primarily include buttons, 
links, tabs, text fields, select options, and other 
related interactable UI elements. 
Certain attributes also help in identifying these elements, 
e.g., role=tab, onclick, aria-label, etc. 
This step will typically reduce the set of HTML elements 
on a page by an order of magnitude (e.g., from 
4564 elements down to 371 elements on cabelas.com). 
\Cref{appdx:html_extra} contains the exhaustive list 
of CSS selectors.

\begin{algorithm}[!t]
\small
\label{algorithm}
\caption{Where $\mathsf{string}$ is the HTML element's attribute 
value and $\mathsf{threshold}$ separates noisy strings from 
strings with information. Updating the dictionary of previously 
seen strings is excluded from the below pseudocode for simplicity.}
\begin{algorithmic}[1] \label{alg:complement}
\Function{DetectNoisyString}{$\mathsf{string}, \mathsf{threshold}$}
\If{len(string) > 2 and len(string) < 100}
    \State $\text{num\_guesses} = \text{zxcvbn}\left(\text{string}\right)$
    \State $\log_{2}\left(\text{num\_guesses}\right) / \text{len(string)}$
    \State word = string contains dictionary words
    \If{not word and score > threshold}
        \State
        \Return True
\EndIf
\ElsIf{len(string) >= 100}
    \State
    \Return True
\Else
    \State
    \Return False
\EndIf
\EndFunction
\end{algorithmic}
\end{algorithm}

\textbf{Element String Matching: }
The next step involves limiting the HTML element search 
space by selecting only elements with strings relevant to 
the website's current state. 
For example, if the screenshot response returns 
\textit{``click the ``Camping'' category on the 
navigation bar''}, then: 
\[[``explore", ``camping", ``outdoors", 
``navigation", ``menu"]\] 
will be the set of strings to search through the element list. 
Or. if the response is \textit{``click ``Tents \& Shelters'' 
under the ``Camping'' category''}, then the list of search 
strings may be: \[[``shelters", ``tents", ``camping"].\] 
By limiting the element search space in this manner, 
the list of candidate elements is often reduced by 
another $10\times$ (e.g., from 371 elements to 29 
elements on cabelas.com).

\textbf{HTML Attribute Cleaning: }
Upon filtering for these elements, only a small set of 
HTML attributes are kept --- those typically containing 
useful information relevant to the functionality of 
the element. These are attributes such as aria-label,
role, type, placeholder, name, title, class, href, etc. 
We also preserve important data nested as child elements 
like plaintext or img tags, as these tags sometimes 
contain information relevant to the functionality of 
an element. Finally, we calculate the entropy of each HTML 
element's attributes' names and values, searching for and 
removing highly random strings in element attributes. 
This is accomplished using the \textit{zxcvbn} password 
strength estimation tool \cite{wheeler2016zxcvbn} which 
uses a corpus of common English words and strings. 
Our system also removes very lengthy attribute 
values, resulting from long href or source URLs. 
By filtering out these strings, we can minimize input 
length and denoise our inputs. 
For example, sometimes the class attribute contains random 
hashes generated by frontend frameworks like Material or React. 
An example of the distillation of HTML content can be seen 
below. Further details like CSS selectors and HTML 
attributes, are available in \cref{appdx:html_extra}.

The HTML cleaning approach by Deng \etal
\cite{deng2023mind2web} reduces these page representations 
down from a median of 47k characters to 17k tokens. 
Compared to their approach, we reduce the length of 
website HTML representations from 
37k down to only 1.4k tokens. 
This 33$\times$ reduction of tokens strikes a balance 
between information retention and conciseness. (\Cref{fig:html_tokens})
While this results in the ground truth element to select 
only presiding in 82.64\% of the tested web pages, 
it is more than sufficient in handling most tasks and 
is a justifiable trade-off due to the increased accuracy 
from reducing the candidate list of HTML elements. This CSS selector list can also be updated to support additional elements.

\subsection{Natural Language Component}

Using a combination of HTML parsing and LLM prompting, we 
can construct a representation of the current page/s state. 
This state consists of a high-level user goal, the 
website's base URL and current page context, a screenshot 
of the current page, a list of prior actions, a proposed 
candidate action, and a list of candidate HTML elements 
to interact with. 
These states are formatted as variables within prompts, and 
the prompts used by each component do not necessarily include 
all state variables. Thus, changes in the state of a web page 
can be measured with this state representation. For prompts 
that exceed a language model's context window, the system 
employs batching of the prompt variables. The following 
components use LLMs process the webpage state and serve as 
the execution flow required to perform an action on a site:

\begin{enumerate}
    \item Summarize the current webpage's context.
    \item Process the page screenshot and suggest a candidate 
        action to perform.
    \item Propose the top 15 elements to interact with to 
        achieve the user's goal.
    \item Select the next best action and element combination 
        to perform from the proposed top-15 elements.
    \item Determine the text to type in or the option(s) to 
        select.
    \item Determine whether the selected candidate action 
       and element makes sense to perform.
    \item Determine whether the current state has achieved 
       the user's goal, and thus the program should terminate.
\end{enumerate}


\begin{tcolorbox}[width=3.3in,
    boxsep=0pt,
    left=4pt,
    right=4pt,
    top=4pt]
\begin{center}\large{HTML element filtering \& cleaning 
example}\end{center}
\rule{\linewidth}{0.6pt}\\
\footnotesize
\textbf{Before Processing}
\begin{lstlisting}[style=htmlcssjs]
<li class="hidden" role="menuitem">
    <a id="departmentButton_3074457345616967" href="https://www.cabelas.com/shop/en/bargain-cave-sale-and-clearance" class="departmentButton navBC" aria-haspopup="true" data-toggle="departmentMenu_3074457345616967">
        Bargain Cave
    </a>
    <div ...>
    </div>
</li>
\end{lstlisting}
\rule{\linewidth}{0.6pt}
\textbf{After Processing}
\begin{lstlisting}[style=htmlcssjs]
<li class="hidden" role="menuitem">
    Bargain Cave
</li>
\end{lstlisting}
\rule{\linewidth}{0.6pt}
\end{tcolorbox}%

\begin{figure}[t]
    \centering
    \includegraphics[width=\columnwidth]{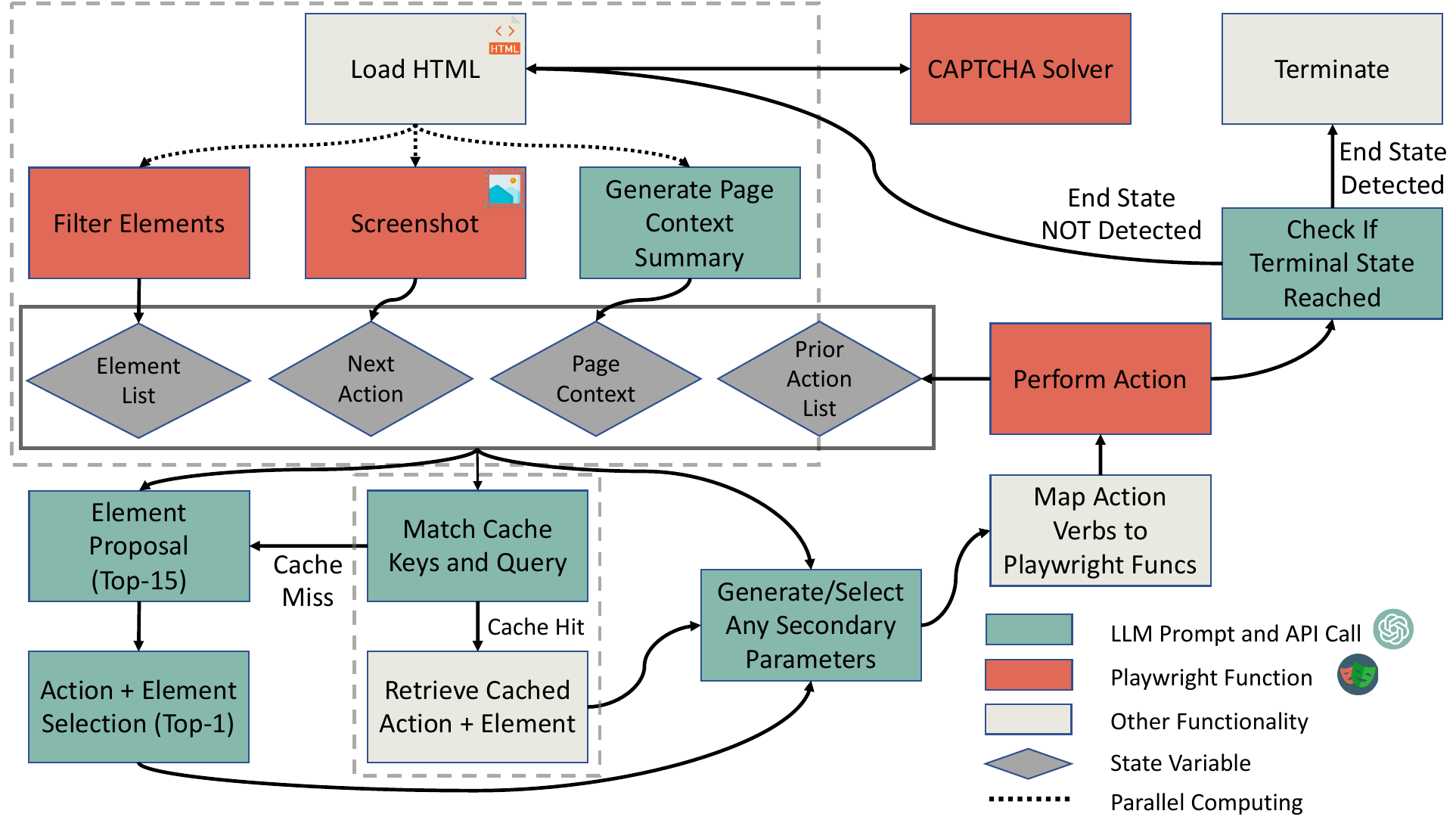}
    \caption{Execution loop for a website and high-level task.}
    \label{fig:execution_loop}
\end{figure}

\Cref{fig:execution_loop} shows a diagram of each of 
these components and how they are integrated into \name.

\textbf{State Representation: }
\name leverages basic prompting techniques like few-shot 
prompting~\cite{logan2021cutting,reynolds2021prompt}, 
chain of thought reasoning~\cite{wei2022chain}, and prompt 
optimization~\cite{yang2023large} to elicit better performance 
from the language models. In particular, the prompt contains 
only the minimum set of variables required for the language 
model to perform its task. The context inputs and response 
outputs are also kept minimal length to ensure quicker 
API response times.
In the cabelas.com example, our system represents a state 
in the text block shown below. The state below
contains information for the language model to 
cater its responses to the current context of the web page. 
Because of the embedded goal, prior actions, and candidate 
elements, the state also contains information for complex 
decision-making, e.g., selecting the next action and element to 
interact with, or determining if the state has completed the 
user's goal. The full prompts used for each component of our 
system are available in \cref{appdx:prompts}.


\textbf{HTML Element Proposal (Top-15): }
Language models at lower temperature settings tend to produce 
outputs that are more deterministic, but simultaneously more 
simplified, conversely, higher temperature results in more 
random outputs. To address this, our system represents HTML 
elements in an indexed list as shown in the 
example state representation above. 
The approach of using indexes to select elements allows for 
more consistency in the outputs of the element 
proposal and selection steps, analogous to the set of marks prompting approach~\cite{yang2023set}. 
An example response looks like: \[ELEMENTS 
[9,1,22,109,84,31,33,77,72,81,117,4,50,54,41]\]


\begin{tcolorbox}[width=3.3in,
                  boxsep=0pt,
                  left=4pt,
                  right=4pt,
                  top=4pt]
\footnotesize
\begin{center}\large{STATE}\end{center}
\rule{\linewidth}{0.6pt}
\textbf{Website:} https://cabelas.com\\
\textbf{Page Context:} Outdoor gear and sporting goods website 
offering hunting, fishing, camping, shooting, and outdoor 
equipment.\\
\textbf{User Goal:} Add a dome tent to my shopping cart.\\
\rule{\linewidth}{0.6pt}
\textbf{PRIOR ACTIONS:}\\
\rule{\linewidth}{0.6pt}
- click the ``Camping'' category in the top menu\\
- click ``Close'' on the popup window with the promotion offer\\
\rule{\linewidth}{0.6pt}
\textbf{NEXT ACTION:}\\
click ``Tents \& Shelters'' under the ``Camping'' category\\
\rule{\linewidth}{0.6pt}
\textbf{CANDIDATE ELEMENTS:}\\
\rule{\linewidth}{0.6pt}
\footnotesize
\begin{lstlisting}[style=htmlcssjs]
(1)
<a class="menulink" role="menuitem" href="/l/tents">
    tents
</a>

(2)
<a id="departmentbutton_3074457345616967301" href="" class="departmentbutton">
    camping
</a>
...
\end{lstlisting}
\rule{\linewidth}{0.6pt}
\end{tcolorbox}

These 15 elements are then used as the list of candidate 
elements to select from in the next step.

\textbf{Element Candidate Selection (Top-1): }
The top element and corresponding action verb are then 
selected from the top 15 elements. The response is 
returned in the format below, and mapped and performed 
via Playwright functions using the mapping defined 
in \cref{table:verb_map}:

\textsc{``action\_verb''} \textit{element\_index}

\textbf{Double-Checking: }
After proposing the top 15 HTML elements, another prompt 
is used to determine whether the proposed candidates 
contain at least 1 element that makes sense given the next 
action state variable. This step is meant to sanity-check 
incorrectly hallucinated element indexes by double-checking 
the LLM response using the top 15 elements' HTML.

\textbf{Text Entry and Option Selection: }
If the selected element and action pair is a 
checkbox/dropdown option (select element) or requires 
typing, the system must generate these secondary 
parameters based on the user's goal. 
This step is used to generate search queries, input fields, 
items to select, etc. The returned response is directly 
used as the text to type into the field, or 
in the case of a checkbox, the index 
of the option is used as the option to select.

\textbf{Terminal State: }
Finally, after each action is selected and performed, 
the system must determine whether the context has reached 
the terminal state in which the user's goal has been achieved. 
Only after this LLM responds with ``yes'' for the end state, 
does the program terminate.

\subsection{Caching Action Sequences}

\name also supports caching previously performed actions to 
reduce runtime and costs. The cache is first keyed by the 
website URL. Subsequent key--value pairs map high-level 
descriptions of actions to the corresponding action verb 
and HTML element. A language model and prompt are used to 
determine the semantic similarity between the stored 
action description (key) and the new action description 
generated by the screenshot response of the current page. 
This step is done such that the only LLM query required for 
each previously executed action is the screenshot response.

\textbf{Cache Indexing: }
The cache is implemented as a dictionary that uses the 
stripped base URLs and the screenshot response containing 
the action descriptions as keys. The selected element 
and action/verb performed on Playwright are stored as the value 
to retrieve. Finally, the cache write/read timestamps 
are stored as metadata for the cache replacement policy. When storing new cached actions, another LLM
component is used to ensure the action description
matches the selected verb and element.

\textbf{Cache Replacement Policy: }
The cache has a maximum number of action description keys of 
100 to reduce issues resulting from language models having 
large input sizes. This is done to keep response times low 
while maintaining consistency and reducing overlap with 
the next action key, as LLMs are known to perform worse on 
reasoning benchmarks with longer inputs. The cache 
supports either LRU or LFU for its replacement policy.



\section{Implementation}

\name was implemented with 1626 lines of code in Python 
and 20 prompts. Its main functionalities rely on 
OpenAI's API, Playwright, BeautifulSoup4, lxml, zxcvbn, 
NLTK's corpus of words, and NopeCHA's API. 
Depending on the component, the system utilizes 4 
different models: GPT-3.5-Turbo, GPT-3.5-Turbo-16k, 
GPT-4-Turbo, and GPT-4-Vision. The system makes use 
primarily of Playwright's Context, Page, and Locator 
classes and their respective functions. 
\Cref{table:verb_map} contains a list of verbs used in 
the LLM prompts and their mapping to Playwright functions.

\textbf{Browser Automation.}
As \name needs to automatically perform tasks on websites, 
it uses the Playwright browser automation framework to 
interface with websites.

Playwright locators retrieve all visible interactable 
elements (e.g., \textless{}a\textgreater{}, 
\textless{}button\textgreater{}, 
\textless{}textarea\textgreater{}, 
\textless{}input\textgreater{}, [class*="ui-slider"], etc.), 
and extracts the outer HTML of each element. 
After the elements and interactions are chosen by 
the NLP component, \name automatically performs the 
actions using Playwright selectors and locators and 
the click, type, select, goto, upload file, 
and enter functions. 
As these actions are implemented, each interaction and 
state is recorded. A screenshot is taken, and, 
optionally, a video recording begins to capture the 
browsing session. Finally, the browser state is stored. 
This includes data, such as network traffic, browser 
cookies, and local browser cache storage. 
The cached actions and elements performed in prior runs 
are retrieved and stored in a JSON file for simplicity.

\textbf{Bypassing Bot Detection.}
To perform activities on behalf of the user, developer, 
or researcher, \name must bypass bot detection techniques,
such as Cloudflare. These techniques typically first 
examine the browser's user-agent header, IP address, browser 
configurations such as headless mode, and other information 
to determine whether the user is likely a bot. 
If these heuristics suggest so, or if the user is 
performing sensitive tasks such as signing into an account, 
these bot-detection services often prompt the user to 
click on a UI element or complete a visual reasoning task 
to prove that they are a human. We bypass
bot detection by using browser configurations 
that mimic normal user configurations. 
In our evaluation, we use the Firefox Nightly browser 
with the UI configuration and a VPN.

\begin{table}[t]
\caption{Verb Mappings to Playwright Actions}
\begin{center}
\scalebox{0.7}{
  \begin{tabular}{c|c}
    \toprule
    {\bf LLM Verb} & {\bf Playwright Function} \\ 
    \midrule
    {\bf click} & Locator.click() \\
    {\bf type\_text} & Locator.fill(text) and Locator.type(text) \\
    {\bf select\_option} & Locator.select\_option(options) \\
    {\bf press\_enter} & Locator.press(`Enter') \\
    {\bf upload\_file} & Locator.set\_input\_files(file) \\
    {\bf visit\_url} & Page.goto(url, wait\_until=`networkidle') \\
    {\bf switch\_tab} & Page.bring\_to\_front() \\
    {\bf close\_tab} & Page.close() \\
    \bottomrule
    \end{tabular}
}
\end{center}
\label{table:verb_map}
\end{table}

\section{Evaluation of \name}
\label{sec:methodology}

\noindent We evaluate \name's efficacy, 
and describe the dataset details and model configurations
in \cref{subsec:dataset}. The setup for each of evaluation 
criteria --- accuracy, runtime, and cost --- is discussed 
in \cref{subsec:eval_method}. Accuracy is measured 
using component per-step accuracy, the system's 
end-to-end correctness, and the system's performance 
running tasks on live websites. Our evaluation results 
are provided in \cref{sec:results}.

\subsection{Dataset Configuration}\label{subsec:dataset}

To evaluate \name's efficacy as a web automation tool, 
we use the Mind2Web dataset~\cite{deng2023mind2web}, which 
consists of 2,350 natural language tasks to perform on 
real websites and over 10,000 recorded actions. 
This dataset was generated using Playwright, and includes 
the actions selected, elements selected, the raw HTML, 
and other metadata. Our system is built to run with 
off-the-shelf models and thus has not been fine-tuned 
{\em a priori} on any datasets.
We use the Mind2Web test set solely to evaluate \name's
performance on various tasks. In our experiments, we
randomly sampled a subset totalling 122 tasks and 621 
actions from the
(test\_domain, test\_site, test\_task) test sets.

\textbf{Language Models Used.}
We utilize and report results with 3 different base models:
GPT-3.5-Turbo, GPT-4-Turbo, and GPT-4-Vision. 
\Cref{table:component_cost_model} in the Appendix provides 
more details on the components and their corresponding models.

\subsection{Evaluation Methodology}
\label{subsec:eval_method}

\textbf{Overall Evaluation (Correctness, Runtime, Cost): }
We evaluated each component in an end-to-end fashion
on subsets of the Mind2Web test dataset which consists 
of 122 tasks and 621 actions.
To ensure the correctness of the web automation sequences, 
we mapped the actions directly with the ground-truth 
actions and elements provided in the Mind2Web dataset. 
One limitation of this dataset is that it is annotated 
with only one correct ground-truth action sequence 
per task, an incorrect assumption.
In reality, completing a task on a website could be 
accomplished in any number of alternative sequences. 
To address this limitation, we manually evaluated another 
subset of 30 tasks from our randomly sampled test set and 
reported the behavior correctness at each action step.


\textbf{Component Per-Step Accuracy: }
Each of the components was evaluated in isolation on larger
200-sample subsets of from the Mind2Web dataset and the 
same samples were used for comparing different models. 
The element proposal samples typically contained 100--200 
elements to select from, whereas the element + action 
selection samples contained 3--5 elements to select from. 
These components were evaluated in isolation to select the 
optimal models to use with each component.

\textbf{System End-to-End Correctness: }
For the end-to-end evaluation, the best-performing models for 
each of the components were integrated into the designed 
system as depicted in \cref{fig:execution_loop}, with 
the primary evaluation focused on the performance of 
the system: element proposal $\xrightarrow{}$ action + 
element selection $\xrightarrow{}$ secondary parameters. 
Thus, the performance of the element + action selection is 
dependent on the element proposal. In this evaluation, 
each sample consisted of a user-defined task that contained 
a set of action steps (typically 5--10 actions). 
The test subset contained 90 tasks across various websites.

\textbf{System Manual Evaluation: }
Due to the limitations of the Mind2Web benchmark dataset,
we conducted a manual evaluation of \name using tasks from
the Mind2Web dataset. The task dates and particular details
were updated to reflect the date of testing. Using 30 tasks,
we annotated each action the system selected, and the task
success rate, as well as any points of failure. 

\textbf{Component Runtime: }
The computation time for each component was recorded in 
the prior end-to-end evaluation for measuring the system's 
typical runtime. \name's LLM prompts are crafted to minimize 
response time and return only the information essential 
to continue operation. We do not include page load times
nor Playwright interaction times due to (1) variability 
and dependence on network connection and (2) this runtime 
is experienced by a normal website user
and/or browser automation tool user.

\textbf{System Cost and Practicality: }
To evaluate the system's cost, we measured the token count 
for each component's inputs and outputs. Each component's 
token usage was mapped to the API costs for their respective 
best-performing models shown in 
\cref{table:per_step_component}. 
\cref{table:component_cost_model} defines the mapping used 
for computing the cost of each component in the system for 
a single pass (element proposal $n=1$), where $n$ denotes the number of 
retries for the element proposal and selection steps.
\section{Evaluation Results} \label{sec:results}

\noindent 
We have conducted the experiments and evaluations described in 
\cref{sec:methodology}, evaluating the following key metrics. 
\begin{itemize}[noitemsep,leftmargin=0.4cm,topsep=5pt]
\item \textbf{System Correctness, or Component 
Per-Step Accuracy:} We evaluated the performance of 5 of the 
8 natural language processing components with various models. 
The action and element selection performs better than
SOTA. For example, the element + action selection step 
in isolation achieves 81.44\% top-1 accuracy using GPT-4. 
We perform an end-to-end 
evaluation of \name's performance in completing tasks with 
the system's components working in tandem. 
The results indicate Mind2Web's ground truth element and 
action are selected in 46.55--48.50\% of the action steps. 
In our manual evaluation of \name running live on websites, 
it achieved a 40\% task completion rate and was able to get 
through an average of 56\% step completion before 
encountering an error. \name detected an end-state 
and terminated correctly in 71\% of tasks.
(\cref{subsec:results_correctness_all})
\item \textbf{System Runtime, Cost, Caching, Practicality: } 
To ensure \name is practical for web automation, we measured 
the token counts and API costs at each action step and 
upon completing a task. \name is shown to be cost-efficient 
for online tasks with an average cost of \$0.18 USD per task
with a median runtime of 8.52 seconds / \$0.028 USD per action. 
Caching reduced this to a median of 4.8 seconds and 
\$0.013 USD per action. 
In a small evaluation of the caching system, a cache hit 
occurred on 49\% of the actions when repeating tasks.
(\cref{subsec:results_cost})
\end{itemize}

\subsection{System Correctness} \label{subsec:results_correctness_all}

\begin{figure}[t]
    \centering
    \includegraphics[width=.98\columnwidth]{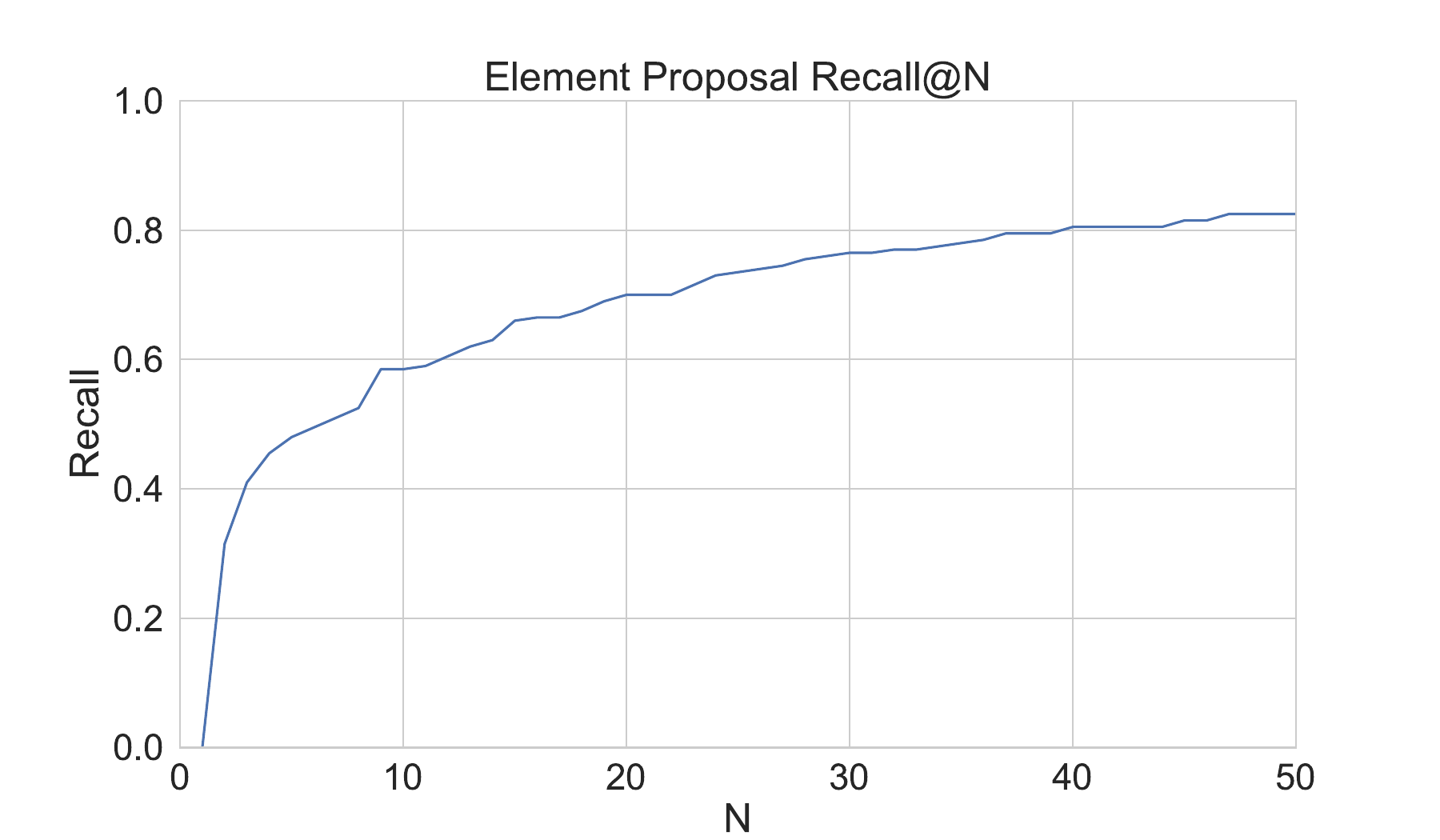}
    \caption{Recall@N score for the element candidate proposal component using the GPT 3.5 Turbo Model.}
    \label{fig:proposal_recall}
\end{figure}

\textbf{Component Per-Step Accuracy: }
\cref{table:per_step_component} demonstrates the 
performance for each component of \name. 
The action and element selection component performs 
consistently. In isolation, our prompting approach 
with GPT-4 achieves an 81.44\% combined accuracy with 
element selection at 83.83\% and action selection at 88.02\% 
(for $n=5$). The element proposal component picks the correct 
element in the top-5 proposals 50\% of the samples, achieving 
a 0.5075 recall@5 score and a 0.7437 recall@25 
(\cref{fig:proposal_recall} portrays the recall up to $n=50$). 
While a higher $n$ yields an improved likelihood of the 
ground truth being in the proposal set, there is a trade-off 
between the number of elements proposed and the performance 
of the element + action selection component. 
The element proposal component was evaluated without reducing
the set of candidate elements using string search.
The fine-tuned GPT-3.5 models perform best for the 
double-checking (0.9192 precision) and end state termination 
(0.7980 F1 score) components, though we use the GPT-4 model
instead of the fine-tuned models in later evaluations for 
simplicity. Finally, for selecting options and typing text, 
all models perform well in this task, with $>83.00\%$ of 
samples matching the ground truth.

\begin{table*}[t]
\begin{center}
\scalebox{0.7}{
  \begin{tabular}{c|c|c|c|c|c|c}
    \toprule
    {\bf Model} & {\bf Element Proposal (Recall@5)} & {\bf @25} & {\bf Element + Action (Top-1)} & {\bf Double Check (Prec)} & {\bf End State Termination (F1)} & {\bf Secondary Param (EM)} \\ 
    \midrule
    {\bf GPT 3.5 Turbo} & \textbf{0.5075} & \textbf{0.7437} & 0.5030 & 0.7525 & 0.4823 & 0.8300 \\
    {\bf GPT 3.5 FT} & -------- & -------- & 0.7305 & \textbf{0.9192} & \textbf{0.7980} & 0.8350 \\
    {\bf GPT 4} & -------- & -------- & \textbf{0.8144} & 0.7900 & 0.6452 & \textbf{0.8650} \\
    \bottomrule
    \end{tabular}
}
\caption{Per-Step Component Results, In Isolation (Mind2Web)}
\end{center}
\label{table:per_step_component}
\end{table*}

\begin{table}[t]
\begin{center}
\scalebox{0.75}{
  \begin{tabular}{c|c|c|c}
    \toprule
    {\bf Success Metric} & {\bf Test Domain} & {\bf Test Task} & {\bf Test Website} \\
    \midrule
    {\bf GT Element in Filtered List} & 61.96\% & 63.98\% & 58.14\% \\
    {\bf Element + Action (Top-1)} & 46.55\% & 48.50\% & 46.70\% \\
    {\bf Text Field Match} & 85.71\% & 90.91\% & 92.59\% \\ 
    \bottomrule
    \end{tabular}
}
\caption{End-to-End System Results (Mind2Web)}
\end{center}
\label{table:end_to_end_system}
\end{table}

\textbf{System End-to-End Step Correctness: }
The system's end-to-end performance on the Mind2Web 
benchmark is lower than the evaluation of 
each component in isolation. This is primarily due to 
(1) errors propagating from the HTML filtering and top 15 
candidate proposal and (2) the limitations of the 
dataset having only 1 ground truth element. 
After filtering for only the list of interactable 
HTML elements, \name retains the ground truth element in 
58.14--63.98\% of steps.
It only proposes the top-1 correctly labeled ground truth 
element and action in 46.55--48.50\% of samples 
(action steps). Text field input generation matches the 
ground truth text input 85.71--92.59\% of the time. 
The components' performance in this end-to-end setting 
are found in \cref{table:end_to_end_system}. 
However, this performance is not representative of 
the system's actual performance. We observed fairly 
frequent occurrences of selected elements that were 
in line with the user's task or elements that were parents/children of the ground truth, but were not considered 
correct according to the ground truth as shown in 
\cref{appdx:incorrect_selection}. The search space of
valid navigation sequences is practically infinite. 
For example, one could use either the search bar or 
the menu navbar to navigate to a product page. 
Thus, we also evaluate more representative 
system's capabilities.

\textbf{System End-to-End Task Correctness (Manual): }
Our tests found that \name was able to perform many tasks, 
including tasks from the Mind2Web dataset, on real websites 
using Playwright. The system performs reliably and 
accurately, especially on popular websites or sites with 
fewer interactable HTML elements. For example, 
\name was able to complete entire tasks, 
consistently perform searches, correctly click
on links and buttons, and fill out forms. The exhaustive 
list of example tasks and the result of running them can
be found in \cref{tab:manual_evaluation}. Of the 30 tasks, 
40\% were successfully completed by our tool. 
71\% of the completed tasks were successfully terminated 
by \name after reaching the end state. For the remaining 
tasks that failed, the most common failure reason was that 
an element required to progress in the task was not in the 
list of interactables or the list of limited elements after 
performing a string search. Following this, the next most 
common issues were that the LLM thought search icons were 
clickable, the end state was detected early, elements were 
hidden, or an issue occurred with the text field generation. 
A couple of failures were also due to the task not having 
account credentials set up and from a website error. Of the 
tasks that failed, the average number of steps completed 
before encountering issues was 2.33. Overall, 6 failures 
resulted from the LLM failing to return a valid sequence, 
6 from HTML filtering, 2 from skipping a step, 2 from website 
errors, and 2 failures resulted from early termination. 
In general, \name consistently completed search or
e-commerce-related tasks, while it struggled to 
complete booking-related tasks.

\begin{table*}[t]
\begin{center}
\scalebox{0.75}{
  \begin{tabular}{c|c|c|c|c|c}
    \toprule
    {\bf Website} & {\bf Natural Language Task Description} & {\bf Completed} & {\bf Progress} & {\bf Termination} & {\bf Failure Reason} \\
    \midrule
    macys.com & \footnotesize{find marriage registry with name JANE DOE} & SUCCESS & 5/5 & SUCCESS & N/A \\ 
    drugs.com & \footnotesize{Show me the page with information about Adderall side effects.} & SUCCESS & 3/3 & SUCCESS & N/A \\
    drugs.com & \footnotesize{Show me the latest FDA alerts.} & SUCCESS & 1/1 & FAILURE & \footnotesize{Late end state termination} \\ 
    tiktok.com & \footnotesize{Show me the tik tok series playlist from brazil} & SUCCESS & 4/4 & SUCCESS & N/A \\ 
    tiktok.com & \footnotesize{Browse the best Australian food songs.} & SUCCESS & 4/4 & SUCCESS & N/A \\ 
    google.com & \footnotesize{Look for a White PlayStation 5 Console and save it} & SUCCESS & 3/3 & SUCCESS & N/A \\ 
    google.com & \footnotesize{Identify Nike Air Women's size 6 cross training shoes that offer free return.} & SUCCESS & 3/3 & SUCCESS & N/A \\ 
    united.com & \footnotesize{Open the baggage fee calculator} & SUCCESS & 6/6 & SUCCESS & N/A \\ 
    imdb.com & \footnotesize{Browse the list of top 250 movies and add the first one to my watchlist.} & SUCCESS & 3/3 & SUCCESS & N/A \\ 
    budget.com & \footnotesize{Find cars that can be picked up at SFO on April 20 and returned on April 27.} & SUCCESS & 6/6 & SUCCESS & N/A \\ 
    healthline.com & \footnotesize{Browse a list of CDB product reviews.} & SUCCESS & 3/3 & SUCCESS & N/A \\ 
    healthline.com & \footnotesize{Find an easy-to-follow evidence-based nutritious vegetarian diet to} & SUCCESS & 7/7 & FAILURE & \footnotesize{Late end state termination} \\ 
     & \footnotesize{lose weight for a diabetic and heart patient, and sign-up to} &  &  &  &  \\ 
     & \footnotesize{to get the results by email buckeye.foodbar@gym.} &  &  &  &  \\
    nba.com & \footnotesize{Find the current league leader in Assists Per Game.} & FAILURE & 3/4 & N/A & \footnotesize{Not in interactable elements list} \\ 
    adoptapet.com & \footnotesize{Find an adult husky near zip code 10019.} & FAILURE & 2/6 & N/A & \footnotesize{Not in interactable elements} \\ 
    adoptapet.com & \footnotesize{Find a shelter for rabbits and small animals within 100 miles of zip 77084.} & FAILURE & 3/6 & N/A & \footnotesize{Skipped a step} \\ 
    accuweather.com & \footnotesize{find the Monthly forecast for Manchester, GB for May} & FAILURE & 3/5 & FAILURE & \footnotesize{Early end state termination} \\
    pinterest.com & \footnotesize{Save on my pins a Halloween costume image.} & FAILURE & 4/9 & N/A & \footnotesize{No account, credentials invalid} \\ 
    ign.com & \footnotesize{Find a walkthrough for the game \"The Legend of Zelda: Breath of the Wild\".} & FAILURE & 3/5 & FAILURE & \footnotesize{Early end state termination} \\ 
    tvguide.com & \footnotesize{Find more films from the director of Smile.} & FAILURE & 1/5 & N/A & \footnotesize{Search icon not clickable} \\ 
    marriott.com & \footnotesize{Start the process of buying a gift card with a beach theme.} & FAILURE & 0/5 & N/A & \footnotesize{Failed to find hidden link} \\  
    redfin.com & \footnotesize{Find a real estate agent job in Atlanta Georgia and apply.} & FAILURE & 3/7 & N/A & \footnotesize{Failed to click hidden link} \\ 
    stubhub.com & \footnotesize{Find \$100 egift card which has Happy Birthday on it for myself and add to cart.} & FAILURE & 2/3 & N/A & \footnotesize{Skipped a step} \\ 
    stubhub.com & \footnotesize{Find a NOFX ticket for a show in Madrid, Spain on May 14 for 1 person.} & FAILURE & 1/7 & N/A & \footnotesize{Search icon not clickable} \\ 
    stubhub.com & \footnotesize{Book 4 tickets in upper tier for any Trevor Noah show in Brisbane, } & FAILURE & 3/7 & N/A & \footnotesize{Double checking failure} \\ 
     & \footnotesize{Australia, before 30 November, view tickets prices with estimated fees.} &  &  &  &  \\
    accuweather.com & \footnotesize{Check the daily forecast in Madison, WI from April 21 - May 1.} & FAILURE & 1/3 & N/A & \footnotesize{Search icon not clickable} \\ 
    foxsports.com & \footnotesize{add WWE superstar ALIYAH to your favorite by following her.} & FAILURE & 3/5 & N/A & \footnotesize{Website error: no search results} \\ 
    rentalcars.com & \footnotesize{Find a large car with lowest price from Apr 28 to May 1 in Zurich.} & FAILURE & 2/6 & N/A & \footnotesize{Not in limited elements} \\ 
    ryanair.com & \footnotesize{Show me tickets for food and drink attractions in Ireland from April 18 to April 19} & FAILURE & 6/8 & N/A & \footnotesize{Not in limited elements list} \\ 
    ryanair.com & \footnotesize{Find a flight from Dresden to anywhere under \$100} & FAILURE & 2/8 & N/A & \footnotesize{Text field generation issue} \\ 
    trip.com & \footnotesize{Find a Hotel in New York for the dates Wed, 17 Apr to Thu, 18 Apr} & FAILURE & 0/8 & N/A & \footnotesize{Text field generation issue} \\ 
     & \footnotesize{1 room for 2 Adults providing 1 Bed with Breakfast and the rent payable at Hotel} &  &  &  &  \\
     & \footnotesize{with Instant confirmation and Free cancellation} &  &  &  &  \\
    \midrule
    \textbf{Overall} & \multicolumn{1}{|c|}{\textbf{Task Completion Rate: 12/30, 40\%}} & \multicolumn{2}{|c|}{\textbf{Progress: 90/161, 56\%}} & \multicolumn{2}{|c|}{\textbf{End State Success Rate: 10/14, 71\%}} \\ 
    \bottomrule
  \end{tabular}
}
\caption{End-to-End System Results (Manual Evaluation, Mind2Web Tasks). Tasks were completed successfully almost all on the first attempt and all within the second attempt. The system was not trained/validated/tuned on any Mind2Web test set.}
\end{center}
\label{tab:manual_evaluation}
\end{table*}

\subsection{System Runtime, Cost, Practicality} \label{subsec:results_cost}
\begin{figure}[t]
    \centering
    \includegraphics[width=.98\columnwidth]{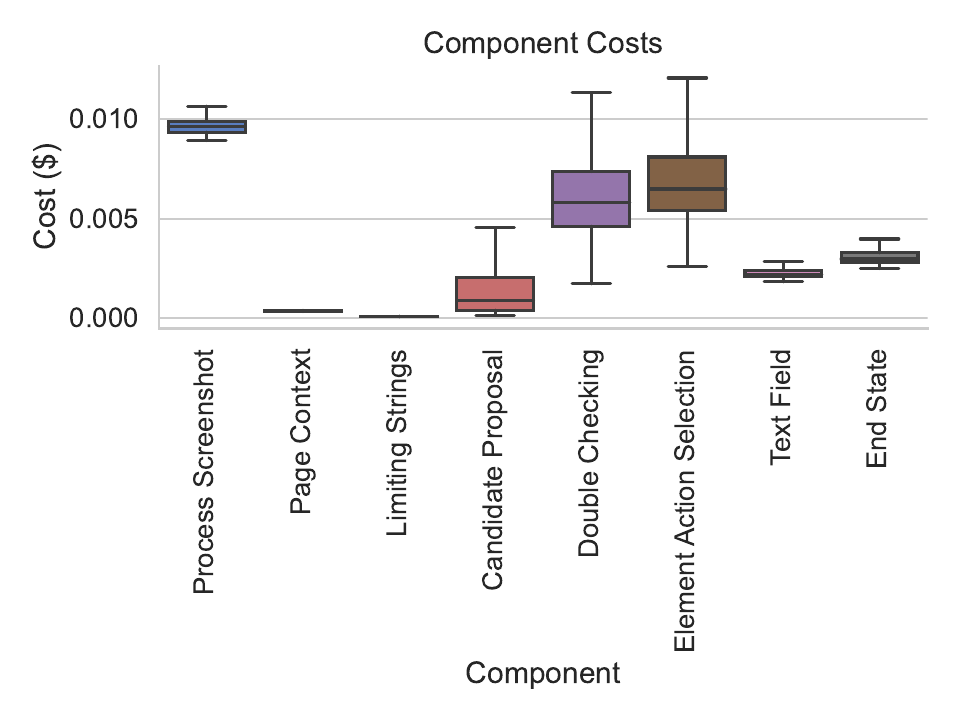}
    \caption{Per-Step API Costs \name}
    \label{fig:component_step_cost}
\end{figure}

\Cref{fig:component_step_cost} presents the distribution of 
API costs per LLM component, across the 122 tasks. 
Most of the incurred costs come from processing the page 
screenshot, checking the top 15 candidate elements, and 
selecting the top 1 element and action pair. 
Each step costs a median of \$0.028 to run.
The system typically has a median runtime of 8.52 to 10.14 
seconds per action step, depending on several variables: 
web page length, number of element proposals, number of 
encountered errors, and whether the action selected is to 
type or select. 
With a cache hit, the system will perform the action with 
a median runtime of 4.8 seconds and a median cost of \$0.013. 
As a reference, it took one of the authors an average of
6.64 seconds (median of 5.98 seconds) to identify 
each webpage element and interact with them by
clicking or pasting text on a subset of the
manually evaluated tasks.

\begin{figure}[t]
    \centering
    \includegraphics[width=.98\columnwidth]{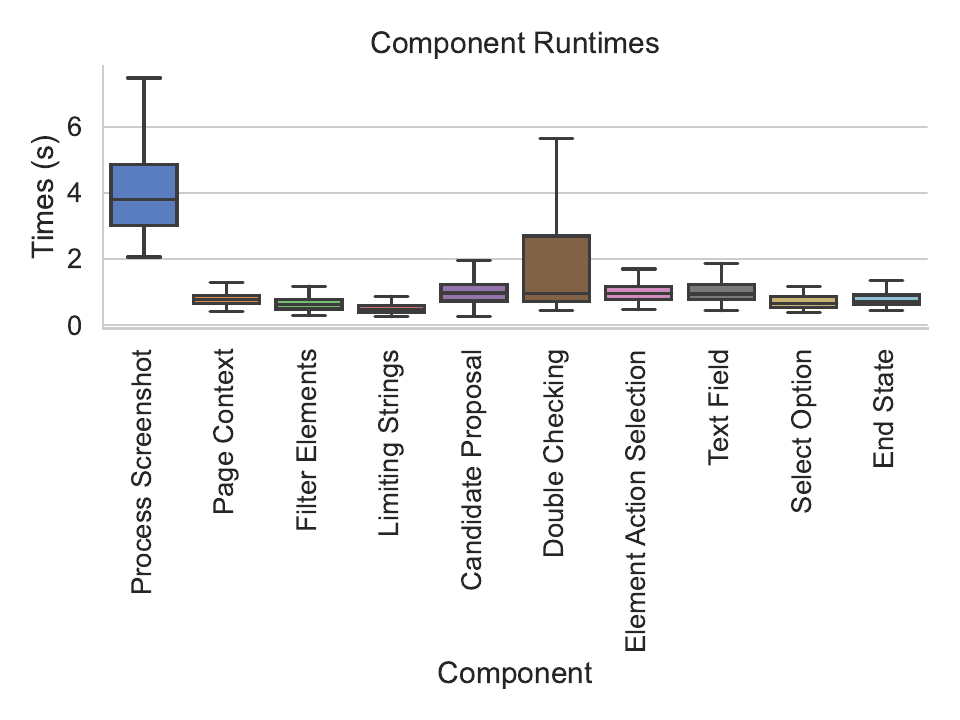}
    \caption{Per-Component Runtime \name}
    \label{fig:runtime}
\end{figure}

\textbf{Caching Performance: }
We evaluated caching performance by running on a set of 
the 5 manually evaluated tasks that successfully completed 
most consistently (macys.com, google.com 2, united.com, 
imdb.com, and healthline.com 1). We ran \name on each task 
10 times after caching the initial set of actions and 
measured the frequency of cache hits. The number of
cache hits for all actions was a median of 4/10 and
an average of 4.9/10. Some of the stored actions
were unnecessary for task completion, and after filtering
these tasks out, the average and median cache hits were
roughly 6/10. Only 2 out of 28 actions were 
duplicate keys in the cache, making most of the keys
unique. All of the cache entries contained correctly
mapped screenshot responses and corresponding
action verbs and elements.

\subsection{Error Analysis}

Much of the current limitations of \name come from its 
first few procedures, the HTML filtering (via CSS 
selectors), the HTML string search limiting, and the 
element proposal step, likely in part due to the 
challenging nature of the task, distilling hundreds 
and thousands of web page elements and proposing the 
top 15 elements. For the more complex combinations of 
web pages and tasks with unintuitive elements 
to interact with, these earlier steps fail.
Even with fine-tuning, preliminary experiments indicate 
performance is only improved by a few percentage points. 
Most of the other errors result from the selection of 
alternative navigation paths different from the ground truth.
To address this, we can further expand the list of CSS 
selectors and regenerate the string search limiters in 
the event of a failure to find an element.
The other most common source of issues, the end state 
detection and termination, can also be improved by adding 
it as a component of the screenshot response generation. 
Alternatively, checking the page screenshot state after 
an end state is detected may reduce false positives 
(early termination). 
Finally, adding a screenshot component for detecting form 
field errors would allow the system to update its state 
with information from website error messages, preventing 
failures resulting from skipping task steps.



\section{Discussion} 
\label{sec:discussion}

\subsection{Design Philosophy}

\name is designed to be easy and practical to use. 
Other related works focus on augmenting AI assistants with 
web browser automation, but end-users will unlikely feel 
comfortable using LLM web automation systems due to the 
numerous safety/security/privacy concerns. We targeted 
\name's design for developers and researchers so that 
it can be easily integrated with any LLM, used to 
automatically generate UI-exercising sequences, 
or used to conduct dynamic analyses of websites. 
Rather than using program synthesis via browser 
automation code generation, \name is more robust 
and deterministic using an element list 
filtering and indexing approach. Finally, it is designed 
to be practical without sacrificing accuracy. 
It is cheaper and more efficient than other 
existing solutions.

\subsection{Ethical Concerns}
\label{subsec:ethics}

Although the datasets do not explicitly include account 
creation, malicious actors may re-purpose the approach of 
\name to automate the creation of fake or bot accounts to 
impersonate people, spread disinformation, drive traffic, 
scalp products, perform DoS attacks, create disingenuous 
reviews, or artificially up-vote content on social media. 
These malicious actors may make the Internet an unfair and 
untrustworthy environment. One option to mitigate these 
threats is to reduce misuse by training the model to 
disallow registration, sign-up, and account creation contexts, 
although this may negatively impact the efficacy of \name.

Additionally, people who use \name as a web assistant may 
provide it with sensitive/authentication credentials. 
This may be risky as sensitive data such as user 
names, passwords, PII, etc., could be unintentionally 
divulged by the language model, inputting them
into incorrect fields.
Thus, these types of web automation and assistant
tools should be limited to experimental developer and 
research use only to mitigate user-end security/privacy risks. 

\textbf{Security/Privacy Concerns: }
We do not recommend using \name or similar LLM web 
automation tools for tasks that require sensitive personal 
information or account access. While \name offers 
advantages like flexibility and runtime/cost-efficiency, 
granting web automation tools access to account and 
settings pages can create security/privacy vulnerabilities. 
An attacker may attempt to hijack web automation tools 
by placing prompt injections 
\cite{liu2023prompt,greshake2023not} 
or adversarial examples~\cite{zou2023universal} on websites,
social media feeds, or comments sections. These attacks
may be capable of altering the system's original task.
Such an attack may redirect the tool to change account 
passwords/2FA, or redirect the tool to visit a malicious website.

\subsection{Limitations and Future Work} 
\label{subsec:limitations}

\name is composed of many components, and although the HTML 
cleaning approach significantly reduces the token count, 
there are other ways to improve its cost-effectiveness. 
For example, substituting the element proposal component with 
a fine-tuned T5 or larger language model may make significant 
improvements in the performance and cost of \name. 
We only conducted low-budget instruction tuning on open-source 
LLMs, so there may be more successful model 
configurations we have not yet explored.
An ideal approach would be training a foundation model on 
the HTML of web pages and fine-tuning the model for the 
specialized components described in this paper, although 
this method is prohibitively expensive. 

While our system supports additional browser actions when 
integrated with Playwright, we have not yet evaluated 
this due to the unavailability of relevant datasets. 

We consider two main approaches that may improve
results: (1) retraining the validity checking component 
to detect and suggest solutions that can be used in 
re-generating a component's outputs, targeting the exact 
error source; (2) training a language model to predict a 
high-level representation of the next action and element 
using the context state. This prediction would be used 
to aid the element proposal and element + action 
selection components.

\section{Conclusion} \label{sec:conclusion}

LLMs have shown a significant promise for automating 
intelligent web navigation and interactions. 
We have proposed,
\name, that enables LLMs to 
truly interact with websites on behalf of users. 
It is a \textit{practical} natural language web automation 
framework augmented with optimizations like HTML 
filtering, caching, and branch prediction. 
It is capable of exercising UIs in a context-aware manner 
and is reliable and accurate, usually selecting an 
optimal action and element sequence for a given user's goal. 
This application of LLMs could enable more complex UI 
exercising and website analyses tools. \name can serve as 
a first step for re-defining how we interact with the 
web.


\bibliographystyle{ACM-Reference-Format}
\bibliography{references}

\appendix
\section{Appendix}

\subsection{Examples of Close/Valid Action and Element Selections Labeled as Incorrect}\label{appdx:incorrect_selection}

\begin{tcolorbox}[width=3.5in,
    boxsep=0pt,
    left=4pt,
    right=4pt,
    top=4pt]
\begin{center}\large{Example ``Incorrect'' Selections}\end{center}
\rule{\linewidth}{0.6pt}\\
\footnotesize
\textbf{Task:} Add a set of wireless headphones with active noise-cancelling feature.
\rule{\linewidth}{0.6pt}\\
\textbf{Ground Truth}
\begin{lstlisting}[style=htmlcssjs]
click <div role="button">wireless headphones</div>
\end{lstlisting}
\rule{\linewidth}{0.6pt}
\textbf{Predicted Selection}
\begin{lstlisting}[style=htmlcssjs]
click <div role="button">wireless headphones noise cancelling</div>
\end{lstlisting}
\rule{\linewidth}{0.6pt}
\textbf{Task:} Find a Hotel in New York City with lowest price ... (April 1st and 2nd).
\rule{\linewidth}{0.6pt}
\textbf{Ground Truth}
\begin{lstlisting}[style=htmlcssjs]
click <a>Hotels</a>
\end{lstlisting}
\rule{\linewidth}{0.6pt}
\textbf{Predicted Selection}
\begin{lstlisting}[style=htmlcssjs]
click <a>Hotels in New York City </a>
\end{lstlisting}
\rule{\linewidth}{0.6pt}
\end{tcolorbox}

\begin{table*}[t]
\caption{Average Action Steps per Subdomain Task}
\begin{center}
\scalebox{0.7}{
  \begin{tabular}{c|c|c|c|c|c|c|c|c|c|c|c|c|c|c|c|c|c}
    \toprule
    \textbf{Subdomain} & Airlines & Auto & Car Rental & Department & Digital & Event & Fashion & Game & General & Ground & Hotel & Movie & Music & Other & Restaurant & Speciality & Sports \\
    \midrule
    \textbf{Actions per Task} & 9.49 & 9.11 & 9.21 & 5.23 & 6.93 & 5.40 & 7.68 & 3.80 & 9.06 & 7.42 & 7.73 & 4.78 & 5.49 & 6.04 & 5.89 & 6.79 & 3.59 \\
    \bottomrule
    \end{tabular}
}
\end{center}
\label{table:subdomain_action_steps}
\end{table*}



\begin{table}[t]
\caption{Component Model Costs}
\begin{center}
\scalebox{0.7}{
  \begin{tabular}{c|c|c|c}
    \toprule
    {\bf Component} & {\bf Best Model} & {\bf Input \$/1M tok} & {\bf Output \$/1M tok} \\ 
    \midrule
    {\bf Screenshot} & GPT 4 Vision & \$10.00 & \$30.00 \\
    {\bf Element Proposal Top-15} & GPT 3.5 Turbo & \$0.50 & \$1.50 \\
    {\bf Cache Matching} & GPT 3.5 Turbo & \$0.50 & \$1.50 \\
    {\bf Tab Management} & GPT 3.5 Turbo & \$0.50 & \$1.50 \\
    {\bf Search Key Generation} & GPT 3.5 Turbo & \$0.50 & \$1.50 \\
    {\bf Page Context} & GPT 3.5 Turbo & \$0.50 & \$1.50 \\
    {\bf Element + Action Top-1} & GPT-4 Turbo & \$10.00 & \$30.00 \\
    {\bf Secondary Parameter} & GPT-4 Turbo & \$10.00 & \$30.00 \\
    {\bf Cache Key Check} & GPT-4 Turbo & \$10.00 & \$30.00 \\
    {\bf Double Check} & GPT-4 Turbo & \$10.00 & \$30.00 \\
    {\bf End State} & GPT-4 Turbo & \$10.00 & \$30.00 \\
    \bottomrule
    \end{tabular}
}
\end{center}
\label{table:component_cost_model}
\end{table}

\begin{table*}[t]
\caption{Dataset Details}
\begin{center}
\scalebox{0.9}{
  \begin{tabular}{c|c|c|c|c|c|c|c}
    \toprule
    {} & {\bf Element Proposal} & {\bf Element + Action} & {\bf Secondary Param} & {\bf High Level} & {\bf Page Context} & {\bf Validity Check} & {\bf End State} \\ 
    \midrule
    {\bf Num Samples} & 7351 & 1650 & 700 & 1500 & 1500 & 1500 & 1700\\
    \bottomrule
    \end{tabular}
}
\end{center}
\label{table:datasets}
\end{table*}

\subsection{CSS Selectors}\label{appdx:html_extra}

\begin{tcolorbox}[width=3.5in,
    boxsep=0pt,
    left=4pt,
    right=4pt,
    top=4pt]
\begin{center}\large{CSS Selectors for Interactable Elements}\end{center}
\rule{\linewidth}{0.6pt}\\
\footnotesize
\begin{verbatim}
button:visible
a:visible
input:visible
select:visible
textarea:visible
[role*="radio"]:visible
[role*="option"]:visible
[role*="checkbox"]:visible
[role*="button"]:visible
[role*="tab"]:visible
[role*="textbox"]:visible
[role*="link"]:visible
[role*="menuitem"]:visible
[role*="menu"]:visible
[role*="tabpanel"]:visible
[role*="combobox"]:visible
[role*="select"]:visible
[class*="radio"]:visible
[class*="option"]:visible
[class*="checkbox"]:visible
[class*="button"]:visible
[class*="textbox"]:visible
[class*="menuitem"]:visible
[class*="menu"]:visible
[class*="tabpanel"]:visible
[class*="combobox"]:visible
[class*="select"]:visible
[class*="suggestion"]:visible
[class*="search-bar"]:visible
[class*="search-result"]:visible
[class*="toggle"]:visible
[onclick]:visible
[href]:visible
[aria-controls]:visible
[aria-label]:visible
[aria-labelledby]:visible
[aria-haspopup]:visible
[aria-owns]:visible
[aria-selected]:visible
\end{verbatim}

\end{tcolorbox}

\subsection{Prompts}\label{appdx:prompts}

This section contains sample prompts used in each of the 7 components 
of \name. These prompts contain both the instruction portion and the
input portion of the prompts. The instructions contain high-level
instructions for the language model to follow, whereas the inputs
contain the current web page state. ``...'' is used to maintain brevity
and typically contains either additional elements to select from,
example states used in few-shot prompting, or are self-explanatory.

\begin{tcolorbox}[width=3.5in,
    boxsep=0pt,
    left=4pt,
    right=4pt,
    top=4pt]
\begin{center}\large{Webpage Context Summary}\end{center}
\rule{\linewidth}{0.6pt}\\
\footnotesize
Take a deep breath. You are an AI assistant made for browsing the web. You will get the text found on a web page. Provide a 1-sentence high level description that summarizes the primary purpose and context of the page. E.g., "Search engine landing page for duckduckgo of the search result for software jobs", "Facebook post creation interface and homepage", "Ecommerce shopping search results page for bubbly soda", "Sign in page with email or phone input for youtube", "Video player home feed with recommendations", "Social media forum detailing todays events, news, community posts", etc.\\\\
\textbf{STATE:}\\
\rule{\linewidth}{0.6pt}\\
Website visited: united.com\\
Page text: ...\\
\rule{\linewidth}{0.6pt}
\end{tcolorbox}

\begin{tcolorbox}[width=3.5in,
    boxsep=0pt,
    left=4pt,
    right=4pt,
    top=4pt]
\begin{center}\large{HTML Element Proposal}\end{center}
\rule{\linewidth}{0.6pt}\\
\footnotesize
Take a deep breath. You are an AI assistant made for browsing the web. You will get a state containing information on a web page, a goal, a list of previously performed actions, and a list of candidate elements. Considering the last actions you took, return the index of the next HTML element to interact with next to achieve your goal followed by a reasoning. Return the single best candidate element. E.g.,"ELEMENT (1)\\\\
\textbf{STATE:}\\
Website visited: united.com\\
Page context: Airline booking and travel information website for United Airlines.\\
Goal: Search the status of flight from Columbus, number 1234 on April 5th, 2023.\\
\rule{\linewidth}{0.6pt}\\
Goal: Select a high speed train ticket with a departure time before 23:00  from Shanghai to Beijing.\\
\rule{\linewidth}{0.6pt}\\\\
\textbf{ACTIONS PERFORMED:}\\
\rule{\linewidth}{0.6pt}\\\\
- None\\
\rule{\linewidth}{0.6pt}\\\\
\textbf{CANDIDATE ELEMENTS:}\\
\rule{\linewidth}{0.6pt}\\
(1) <button class="atm-c-btn--bare" type="button">Close Panel </button>\\
(2) <button type="button\">+</button>\\
(3) <button class="app-components-SearchModal-styles\_\_searchTrigger--ttVhr" type="button"><img alt="" role="presentation">Search for a topic</button>\\
...\\
Based on this state, ELEMENT (9) clicking on the \"Flight status\" tab is the best option.

\end{tcolorbox}

\begin{tcolorbox}[width=3.5in,
    boxsep=0pt,
    left=4pt,
    right=4pt,
    top=4pt]
\begin{center}\large{HTML Element+Action Ranking}\end{center}
\rule{\linewidth}{0.6pt}\\
\footnotesize
Take a deep breath. You are an AI assistant made for browsing the web. You will get a state containing information on a web page, a goal, a list of previously performed actions and a list of candidate elements. Considering the last actions you took, format your response with the best action and element pair to perform next. Return the word "click", "type\_text", "select\_option", "press\_enter", "upload\_file" followed by the index of the element to perform the next action on. \\E.g., "click (1)"\\or\\"type\_and\_enter (4)"\\If none of the candidate elements are appropriate, return just the word "None".\\
\textbf{STATE:}\\
\rule{\linewidth}{0.6pt}\\
Website visited: tiktok.com\\
Page context: This is the homepage of TikTok.com, which offers various tools, showcases, and insights that help marketers and businesses create successful TikTok advertising campaigns by leveraging trends, hashtags, songs, and creators.\\
Goal: Find and show me the analytics for the top trending educational hastag in egypt in the last 120 days that is new to top 10\\
\rule{\linewidth}{0.6pt}\\\\
\textbf{ACTIONS PERFORMED:}\\
\rule{\linewidth}{0.6pt}\\
- click Content navigation with tags and categories.\\
- click Hashtag Navigation Dropdown.\\
- type\_text egypt in Advertising keyword search bar.\\
- click Location option in a dropdown list.\\
- click Timeframe label - ``Last 7 days''.\\
- click Timeframe selection.\\
- click Checkbox for ``new to top 100'' songs.\\
\rule{\linewidth}{0.6pt}\\\\
\textbf{CANDIDATE ELEMENTS:}\\
\rule{\linewidth}{0.6pt}\\
(1) <a class=``link--L6Xda''>\\
  Top Products\\
</a>\\\\
(2) <a class=``hashtagDetailBtn--HjYUJ''>See analytics </a>\\
...
\end{tcolorbox}

\begin{tcolorbox}[width=3.5in,
    boxsep=0pt,
    left=4pt,
    right=4pt,
    top=4pt]
\begin{center}\large{Screenshot Response}\end{center}
\rule{\linewidth}{0.6pt}\\
\footnotesize
You are an AI assistant made for browsing the web. You will get a state containing the desired task to perform on a website, a list of previously performed actions, and a screenshot of the website. Respond with a verb (click, type\_text, select\_option, press\_enter, upload\_file) to perform on an element and a description of the element to interact with next to achieve the task. E.g., "click search button with magnifying glass icon"
\end{tcolorbox}


\begin{tcolorbox}[width=3.5in,
    boxsep=0pt,
    left=4pt,
    right=4pt,
    top=4pt]
\begin{center}\large{Generative Text Inputs}\end{center}
\rule{\linewidth}{0.6pt}\\
\footnotesize
Take a deep breath. You are an AI assistant made for browsing the web. You will get a state containing a goal, a page context, and a high level action. What text would make the most sense to type into the input field? Only return this text. E.g., "New York", "4/5! I thought the restaurant was a great experience", "how to find a job in my neighborhood", "That's awesome, bring me a souvenir!", "french fries", etc.\\\\
\textbf{STATE:}\\
\rule{\linewidth}{0.6pt}\\
Website visited: thumbtack.com\\
Page context: This is the website for Thumbtack, a platform that connects customers with professionals for various services such as home improvement, cleaning, repairs, and more.\\
Goal: find electricians near 10203\\
\textbf{CANDIDATE ACTION:}\\
\rule{\linewidth}{0.6pt}\\
type\_text in Task input field.\\
\rule{\linewidth}{0.6pt}\\
\end{tcolorbox}

\begin{tcolorbox}[width=3.5in,
    boxsep=0pt,
    left=4pt,
    right=4pt,
    top=4pt]
\begin{center}\large{Select Option}\end{center}
\rule{\linewidth}{0.6pt}\\
\footnotesize
Take a deep breath. You are an AI assistant made for browsing the web. You will get a state containing a goal, a page context, a high level action, and a list of options. Return the index of the option to select. E.g., "1", "2", "3", etc.\\\\
\textbf{STATE:}\\
\rule{\linewidth}{0.6pt}\\
Website visited: newegg.com\\
Page context: The page is a product listing page for an online retailer specializing in computer components, electronics, and other tech products.\\
Goal: Build an entry-level pc with an windows 11 64 bit intel i7 CPU with a256gb ssd drive +  4gb ram and adding cheapest component and accessories available.\\
\textbf{CANDIDATE ACTION:}\\
\rule{\linewidth}{0.6pt}\\
select\_option in Featured Items Selector\\
\rule{\linewidth}{0.6pt}\\
\textbf{SELECT OPTIONS:}\\
\rule{\linewidth}{0.6pt}\\
(1) <option value="0">\\
  Featured Items\\
</option>\\\\
(2) <option value="1">\\
  Lowest Price\\
</option>\\\\
(3) <option value="2">\\
  Highest Price\\
</option>\\
...
\rule{\linewidth}{0.6pt}\\
\end{tcolorbox}

\begin{tcolorbox}[width=3.5in,
    boxsep=0pt,
    left=4pt,
    right=4pt,
    top=4pt]
\begin{center}\large{Double Checking}\end{center}
\rule{\linewidth}{0.6pt}\\
\footnotesize
Take a deep breath. You are an AI assistant made for browsing the web. You will get a state containing information on a web page, a goal, and a list of candidate elements. If none of the proposed elements make any sense to interact with given the context and state, respond "No". Otherwise, if even one of the elements makes sense, respond "Yes".\\
\textbf{STATE:}\\
\rule{\linewidth}{0.6pt}\\
Website visited: apartments.com\\
Page context: A web page for finding apartments and homes for rent in various cities, along with tools for managing rentals and advertising properties.\\
Goal: calculate and search rent for a \$6000 monthly income with 30\% rent budget near 90012 area.\\
\rule{\linewidth}{0.6pt}\\\\
\textbf{ACTIONS PERFORMED:}\\
\rule{\linewidth}{0.6pt}\\
None\\
\rule{\linewidth}{0.6pt}\\\\
\textbf{CANDIDATE ELEMENTS: }\\
\rule{\linewidth}{0.6pt}\\
click Menu toggle button.\\
...
\rule{\linewidth}{0.6pt}\\
\end{tcolorbox}

\begin{tcolorbox}[width=3.5in,
    boxsep=0pt,
    left=4pt,
    right=4pt,
    top=4pt]
\begin{center}\large{Check For End State}\end{center}
\rule{\linewidth}{0.6pt}\\
\footnotesize
Take a deep breath. You are an AI assistant made for browsing the web. You will get a state containing the desired task to perform on a website and a list of performed actions. Do the performed actions seem to finish the task? Respond with "Yes" or "No" followed by an explanation.\\
E.g., for this state: ...\\
...\\\\
\textbf{STATE:}\\
\rule{\linewidth}{0.6pt}\\
Website visited: recreation.gov\\
Page context: Recreation.gov is a website that provides tools and tips for discovering outdoor and cultural destinations, offering trip planning, information sharing, and reservations for incredible experiences.\\
Goal: Find campgrounds from April 1st to 4th 2023 that are available at Illinois for 2 adults and 2 kids.\\
\rule{\linewidth}{0.6pt}\\\\
\textbf{ACTIONS PERFORMED:}\\
\rule{\linewidth}{0.6pt}\\
- type\_text Illinois in Search bar.\\
- click Search suggestion: Illinois.\\
- click Accommodation selector.\\
- click Accommodation search dropdown.\\
- click Apply button.\\
- click Check-in date picker.\\
- click Reservation button.\\
- click Date picker input field.\\
- click Selected date on calendar.\\
- click End date input field.\\
- click Calendar day.\\
- click "Search button"\\
- select\_option Price in Search sorting dropdown.\\
- select\_option Available in Search Sorting Dropdown.\\
\rule{\linewidth}{0.6pt}\\
\end{tcolorbox}

\end{document}